\documentclass[conference]{IEEEtran}
\IEEEoverridecommandlockouts
\usepackage{cite}
\usepackage{amsmath,amssymb,amsfonts}
\usepackage{algorithmic}
\usepackage{graphicx}
\usepackage{textcomp}
\usepackage{xcolor}
\usepackage{hyperref}
\usepackage{amsmath}
\usepackage{amssymb}
\usepackage{mathtools}
\usepackage{amsthm}
\usepackage{multirow, makecell}
\usepackage{booktabs}
\usepackage[capitalize,noabbrev]{cleveref}
\usepackage{dblfloatfix}    

\theoremstyle{plain}

\theoremstyle{definition}

\theoremstyle{remark}

\usepackage{xcolor}

\usepackage{array}

\def\BibTeX{{\rm B\kern-.05em{\sc i\kern-.025em b}\kern-.08em
    T\kern-.1667em\lower.7ex\hbox{E}\kern-.125emX}}
\begin{document}

\title{Investigating the Duality of Interpretability and Explainability in Machine Learning}

\author{\IEEEauthorblockN{1\textsuperscript{st} Moncef Garouani}
\IEEEauthorblockA{\textit{IRIT, UMR5505 CNRS} \\
\textit{ Université Toulouse Capitole}\\
Toulouse, France \\
moncef.garouani@irit.fr}
\and
\IEEEauthorblockN{2\textsuperscript{nd} Josiane Mothe}
\IEEEauthorblockA{\textit{IRIT, UMR5505 CNRS} \\
\textit{ Université de Toulouse}\\
Toulouse, France \\
josiane.mothe@irit.fr}
\and
\IEEEauthorblockN{3\textsuperscript{rd} Ayah Barhrhouj}
\IEEEauthorblockA{\textit{LIS, UMR 7020 CNRS} \\
\textit{Aix-Marseille University}\\
Marseille, France \\
ayah.barhrhouj@univ-amu.fr}
\and
\IEEEauthorblockN{4\textsuperscript{th} Julien Aligon}
\IEEEauthorblockA{\textit{IRIT, UMR5505 CNRS} \\
\textit{ Université Toulouse Capitole}\\
Toulouse, France \\
julien.aligon@irit.fr}
}

\maketitle

\begin{abstract}
The rapid evolution of machine learning\,(ML) has led to the widespread adoption of complex ``black box'' models, such as deep neural networks and ensemble methods.  These models exhibit exceptional predictive performance, making them invaluable for critical decision-making across diverse domains within society.  However, their inherently opaque nature raises concerns about \textit{transparency} and \textit{interpretability}, making them untrustworthy decision support systems. To alleviate such a barrier to high-stakes adoption, research community focus has been on developing methods to explain black box models as a means to address the challenges they pose. 
Efforts are focused on \textit{explaining}  these models instead of 
developing ones that are inherently interpretable.
Designing inherently interpretable models from the outset, however, can pave the path towards responsible and beneficial applications in the field of ML.
In this position paper, we clarify the chasm between explaining black boxes and adopting inherently interpretable models. We emphasize the imperative need for model interpretability and, following the purpose of attaining better\,(i.e., more effective or efficient w.r.t. predictive performance) and trustworthy predictors, provide an experimental evaluation of latest \textit{hybrid} learning methods that integrates symbolic knowledge into neural network predictors. 
We demonstrate how interpretable hybrid models could potentially supplant black box ones in different domains.
\end{abstract}


\section{Introduction}
\label{sec:introduction}


In the rapidly evolving field of artificial intelligence, machine learning techniques\,(e.g., Artificial Neural Networks) are among the most widespread tools for high stakes decision-making across diverse domains within society\,\cite{Garouani2022a}. The learning process consists of the model internal hyperparameters tuning in order to mine the useful information buried in the domain data and to maximize the predictive capability\,\cite{Linardatos_2020}. However, despite the impressive predictive power of ML algorithms even in complex scenarios, one of the most intriguing yet challenging aspects is their opacity, intended as their inability to provide intelligible representation of the acquired knowledge\,\cite{Linardatos_2020}. It is non-trivial to forecast what machines will actually learn from data, or whether and how they will grasp general and reusable information for the whole domain making them black box decision support systems\,\cite{Linardatos_2020, simonyanDeepConvolutionalNetworks2014}. As neural networks grow in complexity, they often become so-called ``black-box" models, where the inner workings and decision-making processes become obscure and difficult to understand. This lack of transparency raises concerns, especially in regulated industries such as healthcare, finance, and autonomous systems due to government regulations and guidelines\,\cite{Pitfalls}. 

Current state-of-the-art efforts have made remarkable strides in developing a plethora of methods for opening the black-boxes\,\cite{Ribeiro_2016,Pitfalls}. These solutions encompass a wide range of methods and techniques that shed light on the internal mechanisms of ML algorithms, allowing us to inspect, explain, interpret, and debug their behavior.
Therefore explainability and interpretability are crucial for building trust and ensuring safety in machine learning systems\,\cite{Garouani2022a}.

\textit{Explainability}, often referred to as the capacity to clarify and make understandable the decision-making processes of ML models, plays a crucial role in addressing this opacity. 
In this field, one frequently used approach is the use of visualization techniques, where researchers create visual representations of neural network activations, attention weights, and feature maps\,\cite{simonyanDeepConvolutionalNetworks2014,Linardatos_2020}. These visuals provide insights into how the model processes and transforms input data through its layers, revealing patterns and features influencing decision-making. Features attribution, another method, aims to identify the contribution of each input feature to the model output using techniques like gradient-based attribution or saliency maps\,\cite{ SHAP, simonyanDeepConvolutionalNetworks2014}. This information is crucial in determining which aspects of the input data the model relies on to make its decisions and can help uncover potential biases or vulnerabilities.

\textit{Interpretability}, on the other hand, focuses on the degree to which the model internal mechanisms and features can be understood and linked to real-world concepts\,\cite{Garouani2022a}. In this field, an interesting strategy that that has gained attention is the proactive prevention of ML models from becoming opaque black boxes. This approach involves controlling the training process of the neural networks in a way that incorporates prior knowledge or domain expertise, guiding the model towards learning specific patterns and behaviors\,\cite{KBANN,Magnini_2022}. Along this line, symbolic knowledge injection is the task of letting sub-symbolic predictors acquire symbolic information to impart important constraints and insights, thus influencing the NNs learning behavior\,\cite{Magnini_2022}. This not only helps in developing more interpretable models but also enhances the overall performance and generalization capabilities of the network. Researchers have also explored techniques like model distillation and symbolic knowledge extraction\,\cite{PSyKE}, which involve creating simplified, more interpretable models that approximate the behavior of the original complex NNs, making it easier for humans to comprehend and validate their predictions.

Despite promising results in each of these fields, there is a lack of efficient methods that integrate both \textit{Interpretability} and \textit{Explainability} for AI algorithms. Such an approach would enable the reinforcement of ML models with supplementary knowledge, leveraging the distinct advantages of \textit{Explainability} (actionability, user-friendliness) and \textit{Interpretability} (robustness).
Thus, the objective of this work is to (1) discuss the chasm between explaining black boxes and adopting inherently interpretable models; (2) Clarify the challenges of explaining black box ML models and the development of hybrid interpretable ones; (3) Conduct a comprehensive analytical investigation into the influence of knowledge quality and completeness on the hybrid NN performance; and (4) present a position on the nuanced choice between explainability and interpretability within AI systems. 
Through these objectives, we expect to contribute valuable insights to the ongoing discourse surrounding AI model transparency and understanding.


Accordingly, the remainder of this paper is organized as follows. Section\,\ref{XAI_section} provides an overview of Explainable Artificial Intelligence\,(XAI) and its significance in addressing the black-box nature of AI systems. Section\,\ref{IAI_section} focuses on Interpretable AI, discussing the symbolic knowledge injection in neural networks, its rationale and internal operations. Section\,\ref{empirical_evaluation_section} depicts the methodology to evaluate symbolic knowledge integration into neural network predictors as well as the results. Additionally, we showcase various applications, demonstrating how interpretable hybrid models could potentially supplant black box ones in healthcare and economy domains. Following this, Section,\ref{position_section} is dedicated to presenting our position on the choice between explainability and interpretability within AI systems. Finally, Section\,\ref{conclusion} concludes the paper by providing final thoughts.

\section{Explainable Artificial Intelligence}
\label{XAI_section}
The field of Explainable Artificial Intelligence\,\cite{Linardatos_2020} has seen significant growth in recent years, driven by the proliferation of complex black-box models. Machine learning algorithms have become integral to various aspects of society, often contributing to life-impacting decisions\,\cite{Garouani2022a}. While some applications of ML may not necessitate users to comprehend the inner workings of these systems, in many application scenarios, it remains crucial for human operators to grasp the foundational models\,\cite{Linardatos_2020}. 
A case in point is the regulated industries like healthcare, finance, and criminal justice, where stakeholders are private to employ ML systems that not furnish model explanations due to government regulations and guidelines\,\cite{Bharati_2023}. Furthermore, research has unveiled a substantial connection between comprehension and trust in ML systems\,\cite{Garouani_ideal}.

Researchers have taken various approaches to \textit{open up} black-box models by developing tools that provide explanations. Unlike inherent interpretability found in low-complexity models, these techniques are post-hoc methods applied to pre-trained models. Some of the prominent post-hoc explainability methods include SHAP\,\cite{SHAP}, LIME\,\cite{Ribeiro_2016}, and ANCHORS\,\cite{Anchors}. Model explainability can be categorized into two types\,: global explainability and local explainability. Global explainability enables users to understand the model based on its overall structure, while local explainability focuses on explaining a specific decision made by the model for a given instance or model output. However, despite rapid advancements in XAI, there are still significant gaps that need to be addressed to generalize XAI approaches. Current major XAI methodologies are typically applicable to specific types of data and models, often requiring the pre-configuration of input parameters that are not easily implemented by non-experts(see table\,\ref{tab:3}).

\begin{table}[t]
	\caption[Properties of some local post-hoc state of the art tools.]{Properties of some local post-hoc XAI state of the art tools. \textit{Level} is the interpretability coverage: local or global. \textit{Dependency} specifies particular inputs type. }
	\label{tab:3}
	\centering
	
	\begin{tabular}{lcccc}
		\hline
		\noalign{\smallskip}
		\multirow{2}{*}{XAI method} & \multicolumn{2}{c}{Level} & \multicolumn{2}{c}{Dependency}  \\
		\cmidrule(r){2-3}\cmidrule(r){4-5}                      
		& {Local}   & {Global}      & {Data}    & {Model}     	                  \\
		\noalign{\smallskip}\hline\noalign{\smallskip}
		
		LIME\,\cite{Ribeiro_2016}	& $\bullet$ & $\circ$       & $\bullet$         & $\circ$                          \\
		SHAP\,\cite{SHAP}       & $\bullet$ & $\bullet$     & $\circ$   & $\circ$                      \\ 
				ANCHORS\,\cite{Anchors}    & $\bullet$ & $\circ$       & $\bullet$ & $\circ$                       \\
Saliency\,\cite{simonyanDeepConvolutionalNetworks2014}   & $\bullet$ & $\circ$       & $\bullet$ & $\bullet$                    \\

		\noalign{\smallskip}\hline                      
	\end{tabular}%
\end{table}

One such method called LIME\,(Local Interpretable Model-Agnostic Explanations) is used to explain the importance of features by generating a linear surrogate model based on the output of a data sample\,\cite{Ribeiro_2016}. SHapley Additive exPlanations or SHAP\,\cite{SHAP}  method is a prominent technique within the realm of XAI. Rooted in cooperative game theory, SHAP values provide a measure of the importance of each feature in the prediction of a machine learning model. It aims to assign a fair distribution of contributions from each feature to the final model output, thereby providing insights into the individual impact of input features on predictions.  ANCHORS\,\cite{Anchors}, another technique that focus on identifying influential input areas to establish decision rules that “anchors” the prediction sufficiently. 
Another type of XAI method, known as Saliency\,\cite{simonyanDeepConvolutionalNetworks2014}, constructs visual representations highlighting the importance of features by masking aspects of each sample based on the model perception of the input data. Unlike LIME and ANCHORS, Saliency is specific to artificial neural networks\,(models-specific). 

 Some XAI methods offer low-abstraction capabilities, such as visualizing convolutional filters or illustrating data flow through computational graphs\,\cite{simonyanDeepConvolutionalNetworks2014}. These methods are particularly beneficial for model developers seeking to enhance their models using low-abstraction XAI as a quality metric. Nonetheless, despite these endeavors, concerns have been raised about the practicality of these methods, especially when applied in real-world settings or high-stakes public policy contexts\,\cite{Bharati_2023}. Critics argue that methods like LIME and SHAP are primarily designed to provide local interpretability, offering insights into how a model operates for a specific input by ``approximating'' the black-box model they are attempting to explain, failing to fully capture its underlying nature. Furthermore, such methods may be susceptible to adversarial attacks\,\cite{Pitfalls}. Adversarial attacks involve manipulating input data to deceive the model decision-making process while retaining human imperceptibility. This poses a challenge for XAI, as interpretability features can be exploited or misled, compromising the trustworthiness of the explanations provided. For instance, in image classification, slight alterations to input images, often imperceptible to humans, can lead an XAI system to produce misleading explanations or fail to accurately highlight relevant features\,\cite{baniecki2023adversarial}.

\subsection{Challenges in Explainable Artificial Intelligence}
Numerous methods for explaining machine learning models have emerged through collaborative efforts between academia and industry. Nonetheless, a number of persistent challenges have not received the necessary attention, impeding the broad adoption of explainability techniques\,\cite{Rudin_2019}. We outline here some pitfalls and challenges of explainable ML from an academic and industrial standpoint.

\noindent
\textbf{Challenge 1. Scarcity of Quantitative Evaluation Metrics}

One significant research challenge involves the absence of comprehensive quantitative evaluation metrics for various explainable ML techniques\,\cite{XAI_Metrics}. These techniques offer diverse forms of explanations, yet there is a lack of standardized quantitative measures to facilitate their comparison. This challenge becomes particularly pronounced when dealing with subtle variations of techniques that offer similar functionalities. Consider, for instance, the case of LIME\,\cite{Ribeiro_2016}, an explainability technique that furnishes explanations through feature attributions. A close relative of LIME is xLIME\,\cite{Pitfalls}, which follows a slightly divergent strategy in generating explanations. When an end-user is presented with explanations from both LIME and xLIME, the current state of the art lacks an effective means to quantitatively determine the relative utility of each technique in distinct use cases. Although some research efforts have been directed towards assessing their faithfulness\,\cite{Sippy2020Data,DOUMARD2023102162}, no concrete method exists to establish their relative superiority. Similar challenges extend to other techniques, such as SHAP\,\cite{SHAP} and its various iterations\,\cite{pmlr-v119-sundararajan20b}.

The taxonomy of explainability evaluation, as defined by Doshi-Velez and Kim\,\cite{08608}, encompasses three primary approaches. Two of these approaches involve human judgment, necessitating individuals to assess the explanatory value of techniques within downstream tasks. However, this avenue might not be tenable in all scenarios due to resource constraints or limited expertise. The third approach opts instead for the definition of a proxy task followed by simulation of human behavior. However, none of these methods provides quantitative measurements that can be used to compare different techniques.

\noindent
\textbf{Challenge 2. Scalability Issues}

The existing open-source implementations of well-known explainable ML frameworks such as LIME and SHAP do not scale well. Den Broeck et al.\,\cite{08634} have explored the NP-hardness of SHAP explainability for models like logistic regression and neural networks featuring sigmoid activation. Most of the current XAI tools are not designed to elucidate models constructed using distributed systems like PySpark\,\footnote{\url{https://spark.apache.org/docs/latest/api/python/index.html}}. This limitation significantly restricts their applicability within organizations dealing with substantial data volumes, such as the financial and healthcare sectors, where the demand for explainable ML is particularly pronounced. This constraint applies to both local and global explanations.
For local explanations, the endeavor to provide end-users with low-latency implementations of explainable ML techniques has proven to be arduous, leaving many organizations unable to embrace this capability\,\cite{Pitfalls}. Various prevalent global explanation methods necessitate the acquisition of local explanations for all data points, followed by the application of either an aggregation technique or the identification of the most representative data points to establish a comprehensive overview\,\cite{Ribeiro_2016}. However, executing instance-wise local explainability techniques on extensive datasets proves to be time-intensive and computationally demanding, rendering it non-scalable\,\cite{08634}.

\noindent
\textbf{Challenge 3. Actionability Gap in Explanations}

A prominent challenge in the field of explainable ML research revolves around addressing the question of defining actionability within provided explanations\,\cite{Pitfalls}. Presently, much of the research in the field addresses a single query\,: ``\textit{Why was a specific prediction generated?}". However, a clear gap exists in tackling queries that emerge once a stakeholder has received explanations for a given prediction. For instance, consider a scenario where an object is identified as a car, but the explanation highlights distinct sides of the object. How can the model be adjusted to rectify this divergence in understanding?  In situations where a model prioritizes an atypical feature, such as the driver, when identifying trucks, how can the model be guided to focus on more general characteristics and disregard such unconventional indicators? An individual with a good  credit history, yet being classified into a high-risk group, the explanation being their high credit score. How can the model be educated to align with the intuitive principle of assigning low-risk labels to such creditworthy individuals?

Within industry contexts, explainability promises a step towards confident decision-making. However, existing explanations generated by prevalent techniques lack practical recommendations, leaving both the model developer and end-users uninformed about actionable steps. Although counterfactual explanations\,\cite{Chou_2022}  have shown potential by offering actionable insights to end-users, they still fall short of addressing the needs of model developers. An additional limitation lies in the static nature of current explainability techniques\,\cite{Pitfalls}. Ideally, model developers should be able to engage interactively with the explainability method, receiving explanations alongside actionable suggestions for effecting necessary adjustments. This collaborative process should persist iteratively, enabling model refinement through a cycle of explanation, suggestion, action, and continuous improvement.


In conclusion, while Explainable Artificial Intelligence has made significant progress in improving transparency and understanding of black box models, challenges such as the actionability gap in explanations and scalability issues persist. Bridging the gap between providing explanations and ensuring those explanations are actionable remains a critical concern, as users must be able to comprehend and act upon the insights offered by AI systems. Furthermore, the scalability issue poses a significant barrier in deploying XAI on a larger scale, limiting its widespread adoption and integration into real-world applications. Recognizing such challenges underscores the importance of complementing XAI efforts with a focus on inherent interpretability of AI models. Explainability of AI may provide an additional layer of transparency by creating models that are inherently understandable, aiding in both the explanation and comprehension of AI-driven decisions. By addressing both explainability and interpretability, we can strive for a more comprehensive and effective approach to building trustworthy AI systems. In the subsequent section, we focus on the principles and applications of Interpretable AI, highlighting its role as a valuable complement to the advancements in explainability achieved through XAI.

\section{Interpretable Artificial Intelligence}
\label{IAI_section}

 Explainable AI and interpretable AI\,(IAI) represent distinct concepts. Despite occasional interchangeable use in some research studies, the literature reveals varied differentiation and interconnections between these two concepts. XAI outlines \textit{why} the choice was made but not how the decision was reached. The term IAI outlines \textit{how} the choice was made but not why the criteria used were reasonable\,\cite{Vishwarupe_2022}. 
 Explainability denotes the capacity for a ML model and its outcomes to be presented in a comprehensible manner for humans. Explainability explores the reasoning behind a decision, elucidating the choice made without delving into the exact procedural specifics. On the other hand, the interpretability of ML enables users to comprehend the results of the learning models by revealing the rationale for its decisions\,\cite{Vishwarupe_2022}. Gilpin \textit{et al.},\,\cite{Gilpin2018ExplainingEA}, states that both interpretability and fidelity are integral components of explainability.  They stated that a valuable explanation must not only be intelligible to humans\,(interpretability) but also accurately depict the model's behavior across its entire feature space\,(fidelity). Interpretability becomes imperative to facilitate the sociocognitive facets of explainability, while fidelity aids in validating other model prerequisites or discovering novel explanations. In other words, the fidelity of an explanation determines the precision with which the model behavior is elucidated. Thus, an explanation achieves explainability when it is easily comprehensible to humans and effectively elucidates the model operations.

Sub-symbolic predictors, such as neural networks, have gained widespread usage for extracting valuable insights from data, yet their lack of transparency often earns them the label of ``black boxes". Addressing this challenge, the XAI and IAI communities are actively developing strategies to unbox the black-box, exploring the internal mechanisms of NNs for purposes such as inspection and debugging. One prevalent method involves symbolic knowledge injection and extraction, as seen in approaches like PySKI\,\cite{Magnini_2022} and PySKE\,\cite{PSyKE}.

An alternate strategy for managing the opacity of neural predictors is to prevent them from becoming black-box systems. This can be achieved by controlling the training process of neural networks to enable designers to shape the acquired knowledge directly dictating what should be learned and avoided. Referred to as Symbolic Knowledge Injection\,(SKI), this approach seeks to provide sub-symbolic predictors with symbolic information, potentially guiding or constraining their behavior\,\cite{Magnini_2022}.
The injection of symbolic knowledge offers benefits for training sub-symbolic predictors, mitigating challenges arising from their opacity. In this regard, SKI overrides the need for explicit explanations by reducing uncertainty about predictor behavior—instilling confidence that they will consistently adhere to the injected knowledge. 

Conceptually, most existing symbolic knowledge injection techniques adhere to a common workflow\,(explained in\,\cite{Magnini_2022}) , which can be brieﬂy outlined as follows\,: (i) identify a suitable predictor for the given machine learning task; (ii) generate symbolic knowledge to illuminate specific scenarios or notable situations; (iii) apply the SKI approach to the chosen predictor and the generated knowledge, creating a novel predictor that encapsulates this knowledge; (iv) subsequently, train the novel predictor using available data, following standard procedures. The functional core of SKI techniques remains interchangeable, although certain methods might prove more or less suitable for distinct classes of ML tasks or problems. However, to the best of our knowledge, practical implementations of SKI algorithms are largely confined to proof-of-concept demonstrations or, in many instances, are not readily accessible.

\subsection{Trade-off between accuracy and interpretability}

Achieving high accuracy often involves utilizing complex black-box models, which raise concerns over their lack of human understanding. This has sparked a debate about the accuracy-explainability trade-off, with one side arguing for an inverse relationship between model accuracy and explainability, favoring black-box models for accuracy, while the other asserts that this trade-off is uncommon in practice, advocating for simpler interpretable models\,\cite{Johansson_2011}. Both perspectives assume certain models are inherently more explainable, like low-depth decision trees and linear regression, while others, such as neural networks and random forests, remain opaque. 
Yet, there has been no formal study of the statistical cost of interpretability in machine learning\,\cite{ Bell_2022}. 

In an attempt to initiate a formal study of these trade-offs, Johansson \textit{et al.}\,\cite{Johansson_2011} investigated the balance between accuracy and interpretability in 16 bio-pharmaceutical classification tasks by applying ensemble models and interpretable ones like decision trees and decision lists. Among ensemble methods, random forest-type models show superior predictive performance, highlighting their robustness. The Chipper algorithm slightly outperforms other interpretable techniques, resulting in relatively simple and transparent models. Findings indicate a modest accuracy disparity between the best ensemble\,(random forests) and the primary transparent model\,(Chipper), averaging below 5\% accuracy and 0.1 AUC difference. While opaque ensembles achieve higher accuracy, adopting an interpretable model incurs a limited predictive penalty, underscoring the context-driven trade-off.

More recently, Andrew Bell et\textit{ al.}\,\cite{Bell_2022} empirically quantified the trade-off between model accuracy and explainability in real-world policy contexts. The authors measure explainability using both objective criteria, like human ability to anticipate model output and identify key features, and subjective perceptions of understanding. Contrary to existing literature, the research reveals that explainability is not solely tied to model type\,(black-box or interpretable) and is more complex. The study demonstrate no direct accuracy-explainability trade-off and no inherent superiority of interpretable models in terms of explainability. Additionally, the study identifies that providing more information about a model may not necessarily enhance its explainability, and the utility of local explanations, such as SHAP ones, prove most valuable when they substantially differ from global explanations, and the effectiveness of these explanations is heavily reliant on the specific explainability task assigned to the user.


\section{Empirical evaluation of interpretable NNs}
\label{empirical_evaluation_section}
In this section, our objective is to conduct a thorough analytical investigation of the empirical performance of hybrid neural networks. We are particularly interested in understanding their effectiveness when faced with scenarios involving incomplete or erroneous knowledge, both when coupled with high-quality data and when encountering the opposite situation\,(e.i., not enough training data). In this respect, we describe the used methodology, datasets, and classification algorithms. Subsequently, the experimental results are presented and discussed in substantial detail.

\subsection{Methodology}
We first train a DT algorithm to extract theoretical rules from the training data to be incorporated into the NN model. 
In particular, we rely on accuracy as the preferred metric for both predictive performance and fidelity—where the former measures how good the classifier or the corresponding extracted rules are in classifying data instances in absolute terms, while the latter measures the adherence of the extracted rules. 

To make our experiments fair, we relied on a 10-fold cross validation strategy with randomly chosen training records. Then, using the \texttt{PySKI} tool\,\cite{Magnini_2022}, we conducted the symbolic rules injection into a 3-layers fully connected NN with random weights initialization. During training, we added a dropout layer to prevent the network overfitting. Neurons’ activation functions is the rectiﬁed linear unit (RELU), except for the neurons of the last layer that have Softmax. During training, we choose Adam as optimizer, sparse categorical cross entropy as loss function, and 32 as batch size. In total, for each experiment we train predictors for 100 epochs by relying on the \textit{Accuracy} evaluation metric.

\subsection{\textbf{Case Study 1\,: Socioeconomic prediction}}
The Census Income dataset\,\cite{misc_census_income_20} 
is a widely used dataset in ML and data analysis. It contains demographic and socioeconomic information about 48,842  individuals, collected from the 1994 United States Census. The dataset is commonly employed for classification tasks, particularly to predict whether an individual's income exceeds a certain threshold, such as 50K per year. Thedataset serves as an important benchmark in the field of data analysis and ML, aiding in the development of models for socioeconomic prediction and inequality analysis.

\begin{figure}[h!]
\centering
	\resizebox{1\hsize}{!}{\includegraphics{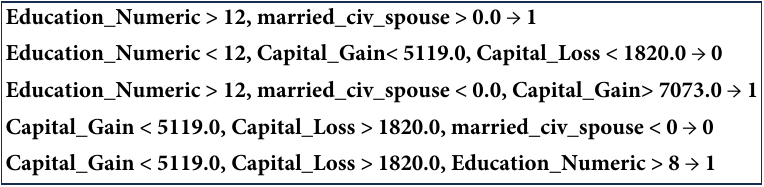}}
	\caption{A sample of the extracted rules from the Census Income dataset.}
	\label{fig:Census_rules} 
\end{figure}

As stated before, the first step is to extract a set of logical rules aimed at predicting whether an individual income exceeds a certain threshold. A sample of the extracted theory\,(rules) is reported in the Figure\,\ref{fig:Census_rules}. Then, we made use of the \texttt{PySKI} tool and methodology\,\cite{Magnini_2022} to inject the extracted prior knowledge into the first layer 
of the 3-layers fully connected NN. Results for all cases are reported in Table \ref{Census_results}.

\vspace{-0.5cm}

\begin{table}[h!]
\caption{Accuracy on the test set for all evaluated scenarios.}
\label{Census_results}
\centering
\begin{tabular}{|>{\centering\arraybackslash}p{1.78cm}|p{3.28cm}|>{\centering\arraybackslash}p{1cm}|c|} 
\hline
\multicolumn{2}{|c|}{Neural network model}                                   & Accuracy~ &  Run-time  \\ 
\hline
\multirow{2}{*}{without injection}                         &  sufficient data & \textbf{85.52}             & 42.06\,s         \\ 
\cline{2-4}
                                          & insufficient data            &   83.62       &  07.53\,s        \\ 
\hline
\multirow{3}{*}{with injection} & sufficient data/correct rules &         \textbf{85.85}           &   69.18\,s       \\ 
\cline{2-4}
                                          & sufficient data/wrong rules            &    84.13      &   74.99\,s       \\ 
\cline{2-4}
                                          & insufficient data/correct rules             &  85.07        & 16.00\,s         \\
\hline
\end{tabular}
\end{table}

The table presents the test set accuracy and run-time for various situations in the context of the predefined NN model, specifically examining the impact of the data and knowledge quality on the model performance. The ``data quality" factor involves two scenarios\,: ``sufficient data" refers to 70\% of data for train and 30\% for test, and ``insufficient data" refers to a test set that is significantly larger than the training set\,(30\% of data for train and 70\% for test). The ``wrong rules” means that we inject erroneous rules to the network.

As expected, in the absence of knowledge injection, the NN model performs best with sufficient training data, achieving an accuracy of 85.52\%. However, in the ``insufficient data" scenario, where the test set is considerably larger than the training set, the accuracy slightly decreases to 83.62\%. This indicates that the model performance is affected by the data quality, showing a minor drop in accuracy when faced with a data scarcity issue.
When knowledge injection is introduced, we observe variations in performance. The NN model with sufficient training data and correct rules achieved an accuracy of 85.85\%, showing that the model performs well when is trained on sufficient data and injected rules are favorable outperforming the same model without prior domain knowledge injection. On the other hand, when erroneous rules are injected into the model while using enough training data, the accuracy decreases to 84.13\%, indicating that erroneous rules negatively impact the model performance.
Interestingly, when the NN model is trained on insufficient data but is provided with correct rules, it still manages to achieve an accuracy of 85.07\%, outperforming the model with the same training data but without prior knowledge of the rules. This suggests that the influence of correct rules can somewhat mitigate the negative impact of data scarcity.
In terms of run-time, the results show that the processing time varies across scenarios. The longest run-time is observed when the NN model is provided with ``sufficient data/wrong rules", taking 74.99 seconds to complete. 
In summary, the results demonstrates that both data quality and theory injection significantly affect the performance of the model. It underscores the importance of considering these factors when deploying NNs, as they can have notable impacts on accuracy and processing time, ultimately influencing the model suitability for specific tasks.

\subsection{\textbf{Case study 2\,: Molecular Biology}}

The primate splice-junction gene sequences\,(PSJGS) dataset\,\cite{PSJGS} consists of 3190 records, each of them represents a sequence of 60 DNA nucleotides, namely adenine, cytosine, guanine, and thymine. Splice junctions are points on a DNA sequence at which ``superfluous" DNA is removed during the process of protein creation in higher organisms. The problem posed in this dataset is to recognize, given a sequence of DNA, the boundaries between exons\,(the DNA sequence retained after splicing) and introns\,(the DNA sequence that are spliced out). 
This dataset has been developed to help evaluate hybrid learning algorithms\,(e.g., KBANN\,\cite{KBANN}, SKI\,\cite{Magnini_2022}) that uses examples to inductively refine preexisting knowledge. The dataset comes with a set of logical rules aimed at classifying DNA sequences provided by human experts. The evaluation of the hybrid model on the dataset is reported in Table\,\ref{PSJGS_results} for all the aforementioned cases.

\begin{table}[h!]
\caption{Accuracy on the test set for all evaluated scenarios.}
\label{PSJGS_results}
\centering
\begin{tabular}{|>{\centering\arraybackslash}p{1.78cm}|p{3.28cm}|>{\centering\arraybackslash}p{1cm}|c|} 
\hline
\multicolumn{2}{|c|}{Neural network model}                                   & Accuracy~ &  Run-time  \\ 
\hline
\multirow{2}{*}{without injection}                         &  sufficient data & \textbf{94.37}             & 38.17\,s         \\ 
\cline{2-4}
                                          & insufficient data            &     91.62     &  27.07\,s        \\ 
\hline
\multirow{3}{*}{with injection} & sufficient data/correct rules &       \textbf{94.65}        &   53.78\,s       \\ 
\cline{2-4}
                                          & sufficient data/wrong rules            &    92.42      &  62.59\,s       \\ 
\cline{2-4}
                                          & insufficient data/correct rules             &   93.04       & 36.85\,s         \\
\hline
\end{tabular}
\end{table}


In this table and in line with the previous case study, the NN model exhibits its highest accuracy of 94.37\% with good training data and slightly lower accuracy of 91.62\% with insufficient training data. These results are consistent with the ones of the previous case study\,: data quality impacts model performance, with a larger training set leading to improved accuracy.
When prior knowledge is injected, the model accuracy remains relatively high. In the "sufficient data/correct rules" situation, the accuracy increases slightly to 94.65\%, indicating that well-crafted rules can enhance model performance. However, when erroneous rules are introduced into the model with sufficient training data, the accuracy drops to 92.42\%, showing that erroneous rules have a negative influence on performance.
In the case of insufficient training data scenario, but with "correct rules", the model obtains an accuracy of 93.04\%, highlighting the compensatory effect of high-quality rules when data availability is limited.

\subsection{Results and discussion}

The findings presented in these case studies underscore the substantial importance of developing hybrid neural networks that incorporate prior domain knowledge to guide their learning process. This highlights a crucial avenue for the development of hybrid NNs that integrate prior domain knowledge or theoretical insights into the learning process. By leveraging established principles and rules from the domain, these hybrid networks could navigate challenges posed by data scarcity or erroneous input more effectively. Integrating such knowledge could potentially guide the network decision-making, enhance its accuracy, and boost its reliability. 

\begin{figure}[h!]
\centering
	\resizebox{1\hsize}{!}{\includegraphics{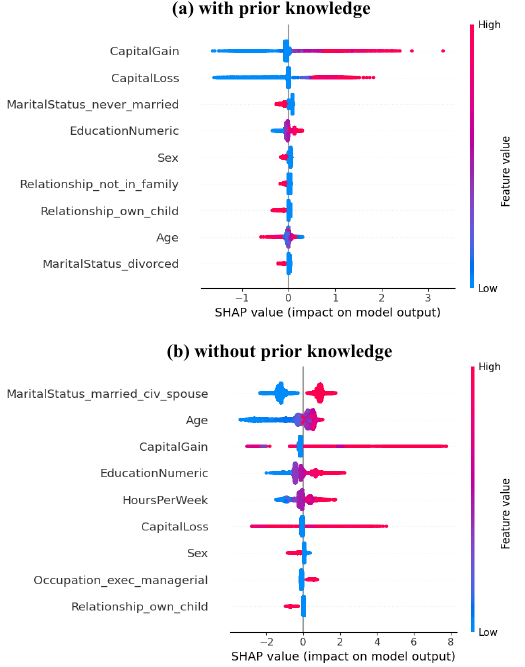}}
	\caption{SHAP beeswarm plot displaying feature importance and impact on the Income prediction\,(case study\,1). 
 }
	\label{fig:shap} 
\end{figure}

To verify the impact and importance of features from the learned model and assess whether domain knowledge improves reliability, we used the SHAP XAI technique\,\cite{SHAP}. SHAP values help understand the influence of each feature on model output, providing insights into how domain-specific rules influence model reliability. Figure \ref{fig:shap} compares feature importance in two models from case study 1—one with prior knowledge and one without. The model with prior knowledge shows higher SHAP values for certain features, indicating their greater influence. Key features like \texttt{CapitalGain, CapitalLoss, MaritalStatus, EducationNumeric} — Cf. figure \ref{fig:Census_rules} drive decision-making, while the model without prior knowledge has more dispersed feature importance, leading to reduced performance and reliability. This supports the claim that embedding domain knowledge enhances model dependability and interpretability.

\section{Explainability vs Interpretability\,: what do we need?}
\label{position_section}
Explainability and interpretability have become some of the hottest topics in recent years with the advent of international regulations on the use of AI. This is not only to validate their performance, but also to build trust between models and users, especially when used in critical decision-making tasks. The choice between explainability and interpretability in AI systems depends heavily on the specific task, the end-users involved, and the regulatory environment. In many scenarios, interpretability and explainability are considered complementary aspects rather than mutually exclusive, as they serve distinct purposes to ensure a comprehensive understanding of the AI system. For instance, a self-driving car system might require interpretability for engineers to optimize the model and explainability for regulators and users to trust its decisions. A more preferred approach involves incorporating both aspects to achieve a balance between providing a comprehensive understanding of the model internal mechanisms and offering transparent justifications for individual predictions. By ensuring the level of interpretability to the specific requirements of the use case, we can develop systems that are not only accurate but also trustworthy and aligned with human values. 
Therefore, the decision between interpretability and explainability should be context-driven, considering the nuances of the application domain and the expectations of end-users.

\vspace{-0.1cm}

\section{Conclusion}
\label{conclusion}

In this paper, we have addressed the critical challenges posed by the adoption of black box ML models. While these models excel in predictive performance, their opacity raises concerns about transparency and interpretability in critical decision-making contexts. The study has highlighted the ongoing efforts to explain these models through visualization techniques, feature attribution, and model distillation, as well as the proactive approach of injecting domain knowledge during training to create more interpretable models.
The findings emphasize the significance complementarity of the explainability and the development of hybrid NNs that incorporate prior domain knowledge, demonstrating their potential to improve accuracy, reliability, and accountability. This complementarity may bridge the gap between theoretical knowledge, data-driven ML and the opening black boxes efforts, offering a promising path toward responsible and beneficial AI applications in domains where trust and transparency are paramount.

\bibliographystyle{IEEEtran}
\bibliography{main_paper}
\end{document}